\documentclass[runningheads]{llncs}
\usepackage[T1]{fontenc}
\usepackage{graphicx}
\usepackage{amsmath}
\usepackage{tabularx}
\usepackage{booktabs}
\usepackage{algorithm}
\usepackage{algpseudocode}
\usepackage{setspace}
\usepackage[style=numeric,backend=biber,sorting=none]{biblatex}
\usepackage{appendix}
\usepackage{booktabs} 
\usepackage{tabularx}
\usepackage{makecell}
\usepackage{hyperref}
\newcommand{\indep}{\perp \!\!\! \perp}

\begin{document}
\title{LUME-DBN: Full Bayesian Learning of DBNs from Incomplete data in Intensive Care}
\author{Federico Pirola\inst{1}\orcidID{0009-0000-7090-4716} \and
Fabio Stella\inst{1,3,4}\orcidID{0000-0002-1394-0507} \and
Marco Grzegorczyk \inst{2}\orcidID{0000-0002-2604-9270}}
\authorrunning{F. Pirola, F. Stella and M. Grzegorczyk}
\titlerunning{LUME-DBN: full Bayesian DBN learning from incomplete data}
\institute{Department of Informatics, Systems and Communication, University of Milano-Bicocca  \\ \email{f.pirola17@campus.unimib.it}  \and
Department of Science and Engineering, University of Groningen
\and
BReCHS – Bicocca Research Centre in Health Services
\and
Bicocca Bioinformatics Biostatistics and Bioimaging Centre - B4, University of Milano-Bicocca, Vedano al Lambro (MB), Italy
}

\maketitle

\begin{abstract}
Dynamic Bayesian networks (DBNs) are increasingly used in healthcare due to their ability to model complex temporal relationships in patient data while maintaining interpretability, an essential feature for clinical decision-making. However, existing approaches to handling missing data in longitudinal clinical datasets are largely derived from static Bayesian networks literature, failing to properly account for the temporal nature of the data. This gap limits the ability to quantify uncertainty over time, which is particularly critical in settings such as intensive care, where understanding the temporal dynamics is fundamental for model trustworthiness and applicability across diverse patient groups. Despite the potential of DBNs, a full Bayesian framework that integrates missing data handling remains underdeveloped. In this work, we propose a novel Gibbs sampling-based method for learning DBNs from incomplete data. Our method treats each missing value as an unknown parameter following a Gaussian distribution. At each iteration, the unobserved values are sampled from their full conditional distributions, allowing for principled imputation and uncertainty estimation. We evaluate our method on both simulated datasets and real-world intensive care data from critically ill patients. Compared to standard model-agnostic techniques such as MICE, our Bayesian approach demonstrates superior reconstruction accuracy and convergence properties. These results highlight the clinical relevance of incorporating full Bayesian inference in temporal models, providing more reliable imputations and offering deeper insight into model behavior. Our approach supports safer and more informed clinical decision-making, particularly in settings where missing data are frequent and potentially impactful.

\end{abstract}

\section{Introduction}
In intensive care units (ICUs), timely and accurate assessment of a patient’s evolving condition is critical for guiding interventions. Bayesian networks (BNs) provide a principled way to represent complex dependencies among physiological variables under uncertainty, and dynamic Bayesian networks (DBNs) extend this capability to temporal domains by explicitly modeling multivariate interactions over time. DBNs have been applied to predict organ failure trajectories \cite{Sandri2014}, forecast physiological changes and mortality risk \cite{chen2022dynamic}, demonstrating their value in integrating heterogeneous ICU time-series data and delivering interpretable decision support in the setting of severe patients care.

However, ICU data are often plagued by missing values, irregular sampling, and measurement biases, which can significantly impair predictive performance and the reliability of the inferred temporal relationships. Traditional imputation or frequentist approaches may fail to fully account for the uncertainty introduced by missingness, potentially leading to biased parameter estimates or convergence to suboptimal solutions. A fully Bayesian approach to missing data imputation in DBNs addresses these challenges by explicitly modeling uncertainty in network structures, parameters and missing values. In this work we propose a Markov Chain Monte Carlo (MCMC) based missing values sampling strategy which naturally accommodates suboptimal local solutions during training, mitigating the risk of convergence to local minima, and efficiently working in case of extensive missingness.

The rest of the manuscript is organized as follows. In Section~\ref{sec:2}, we review the literature on both static and dynamic Bayesian networks learning from incomplete data. Section~\ref{sec:3} introduces a Bayesian formulation of DBNs and the implementation of our method. Section~\ref{sec:4} evaluates convergence and network reconstruction effectiveness in simulated scenarios characterized by different sample sizes and missingness rates. Section~\ref{sec:5} introduces the ICU data employed, describes the preprocessing pipeline adopted to avoid potential systematic missingness and reports the reconstructed networks obtained through our approach. Section~\ref{sec:6} concludes by wrapping up the results obtained in both simulated and real-world settings and illustrates the future directions of research.

\section{Related Works}
\label{sec:2}
The presence of missing values represents a central issue in Artificial Intelligence, significantly influencing model effectiveness and generalizability. Several methods have been developed to handle missing data when learning static BNs from incomplete datasets. A widely used approach is the structural expectation-maximization (SEM) algorithm \cite{ruggieri2020}. In its Hard EM variant, the most practical version, an expectation (E) step imputes missing values using the current model, followed by a maximization (M) step that updates parameters and a structural move (S), iterating until convergence. Despite their effectiveness to work in the context of limited data, such methods always follow the current best local moves and impute missing values based on the best model at each step, which can underestimate uncertainty and lead to convergence to local optima.

Bayesian approaches have recently emerged to overcome these limitations. Recently, \cite{grzegorczyk2023missing} proposed a sampling-based MCMC method that jointly infers structure, parameters, and missing values. Unlike classical score-based or constraint-based structure learning methods, which greedily optimize toward local improvements, these techniques can escape local minima and yield more accurate network reconstructions \cite{grzegorczyk2023missing}.

Despite these progresses, missing data in DBNs is often still handled using model-agnostic methods such as Multiple Imputation by Chained Equations (MICE) or mean imputation, or temporal adaptations of frequentist approaches as SEM\cite{Sandri2014, chen2022dynamic}. 

Since sampling-based MCMC methods have proven effective in static settings, we aim to extend them to DBNs by designing a Gibbs sampling step for recursive missing data imputation. A key insight is that the full conditional distribution (FCD) of missing values in this setting remains tractable, enabling efficient MCMC-based inference, a point we develop further in the methodological section.

\section{Methods}
\label{sec:3}
\subsection{DBNs and Bayesian Linear Regressions}
\setcounter{footnote}{0}
DBNs present a popular and valid approach to model the temporal evolution of multivariate continuous time series.\footnote{For simplicity, throughout the paper we refer to Gaussian DBNs simply as DBNs, although discrete or mixed variants also exist.} They factorize the joint distribution of all variables across time into a product of conditional distributions for each node given its parents in the previous time slice \cite{ghahramani1998learning}. Assuming no instantaneous effects, i.e., variables at time $t$ do not influence other variables at the same time point, there is no need to impose an acyclicity constraint as in static BNs. Under these assumptions, a DBN could be framed as $k$ independent Bayesian linear regression (BLRs) models \cite{grzegorczyk2019modelling}. A DBN is characterized by its structure, consisting of k nodes (variables) \(\{X_1, \dots, X_k\} \) connected through edges, and its parameters, which specify the BLRs. An edge $X_i \rightarrow X_j$ indicates a temporal dependency, namely, variable $X_i$ at time $t-1$ directly influences variable $X_j$ at time $t$.\footnote{For a given edge $X_i \rightarrow X_j$, we define $X_i$ as the parent of $X_j$ and $X_j$ as the child of $X_i$. The set of the incoming edges for a node $X_i$ define its parent set $\pi_{(i)}.$} The parameters, instead, are given in the form of BLRs as in Equation~\ref{eq:linear_model}.

\begin{equation}
x_i^{t} \;=\; \beta_0^{(i)} + \sum_{j: (X^{t-1}_j \in \,\pi_{(i)})} \beta_j^{(i)} x_j^{t-1} + \epsilon^t_{i} \quad  (t = 1, \, \dots , \, T)
\label{eq:linear_model}
\end{equation}

$x_i^t$ is the realization of variable $X_i$ at time $t$, $\beta_0^{(i)}$ is the intercept of the linear regression model and the vector $\beta^{(i)} = [\beta_1^{(i)}, \dots,\beta_{|\pi_{(i)}|}^{(i)}]$ represents the ordered linear coefficients associated with the parents of $X^t_i$ in $\pi_{(i)}$. Finally, $\epsilon^t_{i}$ is a Gaussian-distributed error term with mean 0.

\subsection{Conditional Likelihoods and Prior Distributions}

To learn a DBN from data, we consider a group of \( N \) samples \( \{s_1, \dots, s_N\} \) observed at \( T + 1 \) equally spaced time points \( \{0, 1, \dots, T\} \), where each sample is described by a set of \( k \) variables \( \mathcal{X} = \{X_1, \dots, X_k\} \). This results in a dataset \( \mathcal{D} \) of dimensions \( (N, k, T+1) \), representing realizations of \( \mathcal{X} \) across both the sample and the time domain. Since in DBN we assume the first Markov order, namely $\mathcal{X}^{t+1} \indep \mathcal{X}^{t-1} \,| \,\mathcal{X}^{t}$, we could then rewrite $\mathcal{D}$ in its lagged form $\mathcal{D_L}$ to easily learn the actual model from data. $\mathcal{D_L}$ would then be of dimensions $(N \cdot T, 2k)$ \cite{ghahramani1998learning}. In Equation~\ref{eq:conditional_distribution} we define the conditional Gaussian likelihood for a variable $X^t_i$. 

\begin{equation}
X^{t}_i \;\big|\; (\mathcal{X}^{t-1}, \beta^{(i)}, \sigma^{2(i)}, \pi_{(i)})
\;\sim\;
\mathcal{N}\!\left(
\mathcal{X}^{t-1}_{[\pi_{(i)}]} \, \beta^{(i)},
\sigma^{2(i)} I
\right).
\label{eq:conditional_distribution}
\end{equation}

The regression coefficients $\beta^{(i)}$ quantify the linear effect of parent variables $\mathcal{X}_{[\pi_{(i)}]}$ at time $t-1$ on $X_i^t$, while $\pi_{(i)}$ defines the current parent set for variable $X_i$ at time $t$, and the noise parameter ${\sigma^2}^{(i)}$ capture unexplained variability.

Since we are working in a full Bayesian framework, all the parameters have their own prior distribution and we impose conjugate priors. The regression coefficients follow a Gaussian prior $\beta^{(i)} \sim N(\mu_{(i)}, {\sigma^2}^{(i)} {\delta^2}^{(i)} I)$, where ${\delta^2}^{(i)}$ express uncertainty over the parameters $\beta^{(i)}$. The noise and uncertainty terms are assumed to follow Inverse-Gamma distributions, ${\sigma^2}^{(i)} \sim \text{Inv-Gamma}(a, b)$ and ${\delta^2}^{(i)} \sim \text{Inv-Gamma}(\alpha_{\delta}, \beta_{\delta})$ respectively. Finally, a Poisson prior $|\pi_{(i)}| \sim \text{Poisson}(\lambda) $ over the parent set size is set, encouraging simpler network structures with fewer parents, controlling model complexity. Depending on the application a fan-in restriction could be imposed too, limiting the maximum number of parents per variable. The detailed derivations of the closed-form solutions for the parameters FCDs and the marginal likelihoods can be found in \cite{grzegorczyk2019modelling}.

\subsection{A Gibbs Step for Missing Data Imputation}
\label{sec:3.3}

We design a unified imputation step that updates missing values jointly across all regression models in which they appear. We adopt a Gibbs move, sampling from the full conditional distributions (FCDs) of the missing values. Here for a matter of simplicity we assume uniform priors over the missing values domain, though Gaussian priors would also be possible. Then, a missing value for a variable \(X_i\) at time $t$ contributes to:  
\begin{itemize}
    \item Its own conditional likelihood: \(P(X_i^t \mid \pi_{(i)})\)
    \item The conditional likelihoods of all its children ($X_j^{t+1}$): \(P(X_j^{t+1} \mid \pi_{(j)})\)
\end{itemize}

For the local Markov property of a DBN, all nodes are conditionally independent to their non-descendants given their parents \cite{ghahramani1998learning}. Indeed, for the property of conditional Gaussian distribution, the product of independent conditional Gaussians is also a Gaussian \cite{bishop2006pattern}. It follows that the FCD of each missing value is a univariate Gaussian distribution, thus we can efficiently sample the missing values, enabling a robust Gibbs sampling procedure. 

In Appendix~\ref{app:A} we show the derivation of the missing values' FCDs in an incremental way. We start out by computing the FCD of single missing values in three simple cases and then we illustrate how the derivation of the missing's FCD extends to the general case with more complex Markov blankets. \footnote{The \emph{Markov blanket} of a variable $X_i$ is the set of its parents, its children, and variables sharing children with it. Conditioned on its Markov blanket, $X_i$ is independent from all other variables in the network.}

In Equation~\ref{eq:conditional_missing} we report the FCD for a general missing value ${x_i^t}_{[MIS]}$ over a random variable $X_i$ at a certain time $t > 0$. The FCD is a univariate Gaussian distribution with a closed form whose structure remains unchanged regardless the size of its Markov blanket. The noise term ${\sigma^2}^*$
is a weighted combination of the precisions ($1/{\sigma^2}$) from the conditional likelihood of $X^t_i$ and from the likelihoods of the children $X^{t+1}_j$. The mean term $\mu^*$ is determined by the conditional mean $\mu^t_i$, the isolated contribution $\beta^{(j)}_i$ of $X^t_i$ adjusted by the other parents contributions ${\mu^{(j)}_{\{-i\}}}^{(t+1)}$, and the observed values of the children $x^{t+1}_j$.

\begin{align}
P({x_i^t}_{[MIS]} \mid \cdot)
&= \mathcal{N}\bigl(\mu^*, {\sigma^2}^*\bigr) \label{eq:conditional_missing}\\
\text{where:}\quad
&\begin{aligned}[t]
&{\sigma^2}^* = \left(
\frac{1}{{\sigma^2}^{(i)}} + \displaystyle\sum_{j:\; (X_i^t \in \pi_{(j)})}
\frac{(\beta_i^{(j)})^2}{{\sigma^2}^{(j)}}
\right)^{-1}, \\
&\mu^* = {\sigma^2}^* \cdot \left(
\frac{\mu^t_i}{{\sigma^2}^{(i)}} + \displaystyle\sum_{j:\; (X_i^t \in \pi_{(j)})}
\beta_i^{(j)} \frac{\bigl(x_j^{t+1}- {\mu^{(j)}_{\{-i\}}}^{(t+1)}\bigr)}{{\sigma^2}^{(j)}}
\right), \\
&{\mu^{(j)}_{\{-i\}}}^{(t+1)} = \displaystyle\sum_{\substack{m \neq i: (X^t_m \in \pi_{(j)})}}
x_m^t \beta^{(j)}_m. \nonumber
\end{aligned}
\end{align}

\subsection{Learning DBNs from Incomplete Data}
When data contain missing values, DBN learning procedure can no longer be framed as \(k\) independent BLRs, unless the data are first imputed to obtain a complete dataset, which seems suboptimal. Since any variable may serve as both a response and a covariate, independent imputations per regression would assign inconsistent values to the same missing value and violate the DBN factorization.  

Therefore, we combine the Gibbs sampling step for missing data imputation from Section~\ref{sec:3.3} with the MCMC sampling scheme designed for learning DBNs in case of complete data from \cite{grzegorczyk2019modelling}. We refer to the resulting MCMC scheme for learning DBNs from incomplete data as Latent Uncertainty Modeling via MCMC Estimation in DBNs (LUME-DBN). LUME-DBN is characterized by an iterative procedure sampling the DBN structure, its parameters and the missing values in consecutive steps. 

For the parameter and structural moves, we refer to \cite{grzegorczyk2019modelling}, which illustrates the DBN learning procedure in case of fully observed data. In Appendix~\ref{app:B}, Algorithm~\ref{alg:learn-params} and Algorithm~\ref{alg:learn-parents} provide the detailed Gibbs sampling step for parameter updates sampling from the FCDs and the Metropolis-Hasting step for updating parent sets in each BLR built on marginal likelihoods.

Missing values are, instead, updated jointly across the network once every \(E_M\) epochs, where \(E_M\) is chosen based on application-dependent factors such as missingness rate and convergence time. Conditional on the current structure and parameters, each missing entry is independent across samples, but strictly dependent on the observed trajectory of the corresponding sample. As detailed in Equation~\ref{eq:conditional_missing}, the FCDs are shown to be time-invariant with respect to the DBN's parameters, allowing us to derive efficient index sets and making the imputation step computationally feasible even under high missingness. Algorithm~\ref{alg:learn-incomplete} in Appendix~\ref{app:B} show the entire implementation of LUME-DBN, deriving posterior samples for the structure, the parameters and the missing values.

\subsection{Convergence Diagnostics and DBN Reconstruction Evaluation}
\label{sec:3.5}
After running independent simulations, we obtain a set of posterior samples for the parent sets (binary arc indicators) and the missing values. Since our method focuses on network reconstruction in the context of incomplete data, we want to ensure that both the parent sets, thus the DBN structure, and the missing values converge to the right solutions.

To assess convergence we employ the potential scale reduction factor (PSRF) \cite{gelman1992inference} at each epoch separately for both arc indicators and missing values. A detailed description of PSRF for DBNs arcs is reported in \cite{grzegorczyk2019modelling}. Let $\Phi_e$ denote, for epoch $e$, the proportion of posterior samples with $\mathrm{PSRF} < 1.1$. Convergence is declared when $\Phi_e$ consistently reaches $1$ over successive epochs.  

To evaluate the proposed Gibbs imputation procedure, we compare it in terms of network reconstruction accuracy with two baselines model-agnostic methods for handling missing values in DBN learning. MICE~\cite{Sandri2014} is applied to the original dataset before temporal lagging, regressing each incomplete variable on all the others iteratively using linear models until convergence. Because of the time-lag of all interactions, the original MICE algorithm could result suboptimal in our setting, thus, we implemented \textit{Temporal MICE}, a new variant that uses lagged predictors to predict the missing values.

We discard the first $S$ samples prior to convergence (burn-in) and then thin the chain by retaining only every 5th sample to reduce autocorrelation. Then, we compute the averaged inclusion probabilities, namely the average probabilities of each potential parent in each linear regression model, obtained by averaging the posterior arc indicators across epochs and replicates. The result is a matrix where each row is associated with the probability of the parents for a given outcome variable at time $t$. This matrix is then compared with the original adjacency matrix of the data-generating DBN for each competing method. Model performances are quantified using the area under the precision–recall curve (AUC–PR), where the precision–recall curve plots the trade-off between the proportion of predicted arcs that are correct (precision) and the proportion of true arcs that are recovered (recall) \cite{saito2015precision}. Higher values of AUC–PR indicate better reconstruction accuracy.

\section{Synthetic Data Generation \& Performance Evaluation}
\label{sec:4}
To assess the performance of LUME-DBN, we generate synthetic temporal data from given DBNs. A total of ten independent DBN structures with 10 nodes each are randomly sampled, with each structure constrained to have a maximum of five parents per node. To ensure acyclicity and maintain the faithfulness condition---that is, the compatibility between the model parameters and the underlying graph structure---a random topological ordering of the nodes is imposed during the structure generation process.

Conditional dependencies are modeled using linear Gaussian relationships. For each node, regression coefficients corresponding to its parents are independently drawn from a Uniform distribution over the interval $[0.2, 0.8]$, and the noise variance is fixed at $\sigma^2 = 1$.  Together, the structure, coefficients, and variance parameters fully specify each DBN. 

For each of the ten DBNs, multiple time series are simulated with three different lengths $T = \{50, 100, 200\}$, to evaluate the impact of the number of time frames on structural reconstruction accuracy. \footnote{We also conducted ablation experiments varying sample sizes, maximum number of parents and linear coefficients ranges; the experimental setting reported was chosen to avoid trivially easy or infeasible cases while still covering a broad spectrum of different settings.} To evaluate the robustness of the method to missing data, missing values are introduced randomly at four levels of missingness: $\{10\%, 20\%, 30\%, 40\%\}$. The incomplete datasets at each missingness level are generated by randomly removing data entries according to a Bernoulli distribution with success probability equal to the specified missingness rate.

We assess convergence of both DBN structure and imputed missing values, averaged over 10 simulated DBNs with time series length \(T = 100\). At each epoch $e$, for each missingness level, the convergence diagnostic $\Phi_e$ described in Section~\ref{sec:3.5}, is tracked. Figures~\ref{fig:networkconvergence} and~\ref{fig:missingconvergence} in Appendix~\ref{app:D} shows that structure convergence occurs within 1.5k epochs for all missingness rates, whereas missing values require up to 5k epochs at 40\% missingness. Convergence time increases with missingness, with \(10\%\) missingness resembling the fully observed case and 30--40\% missingness producing more unstable structure learning. Still, in all settings convergence is ultimately achieved within $S = 5k$ epochs, thus we discard the first $S$ posterior samples to ensure the stationarity of the distribution.

We then run five independent simulations of 20k epochs for each experimental condition. Missing values are imputed every \(E_M = 10\) iterations, with initial imputation given by the mean at time \(t\). This value of \(E_M\) guarantees faster converge time compared to an update of the missing values at every single epoch. A fan-in restriction is imposed by randomly initializing the covariate sets with at most five parents. Noise ${\sigma^2}^{(i)}$ and uncertainty ${\delta^2}^{(i)}$ parameters are initialized to 1, linear coefficients $\beta^{(i)}$ to 0, and the Poisson prior parameter is set to \(\lambda = 1\). All experiments were implemented in R, and the full code of LUME-DBN algorithm is publicly available on the \href{https://github.com/federicopirola/LUME-DBN}{GitHub page}.

\begin{figure}[H]
    \centering
    \includegraphics[scale = 0.25]{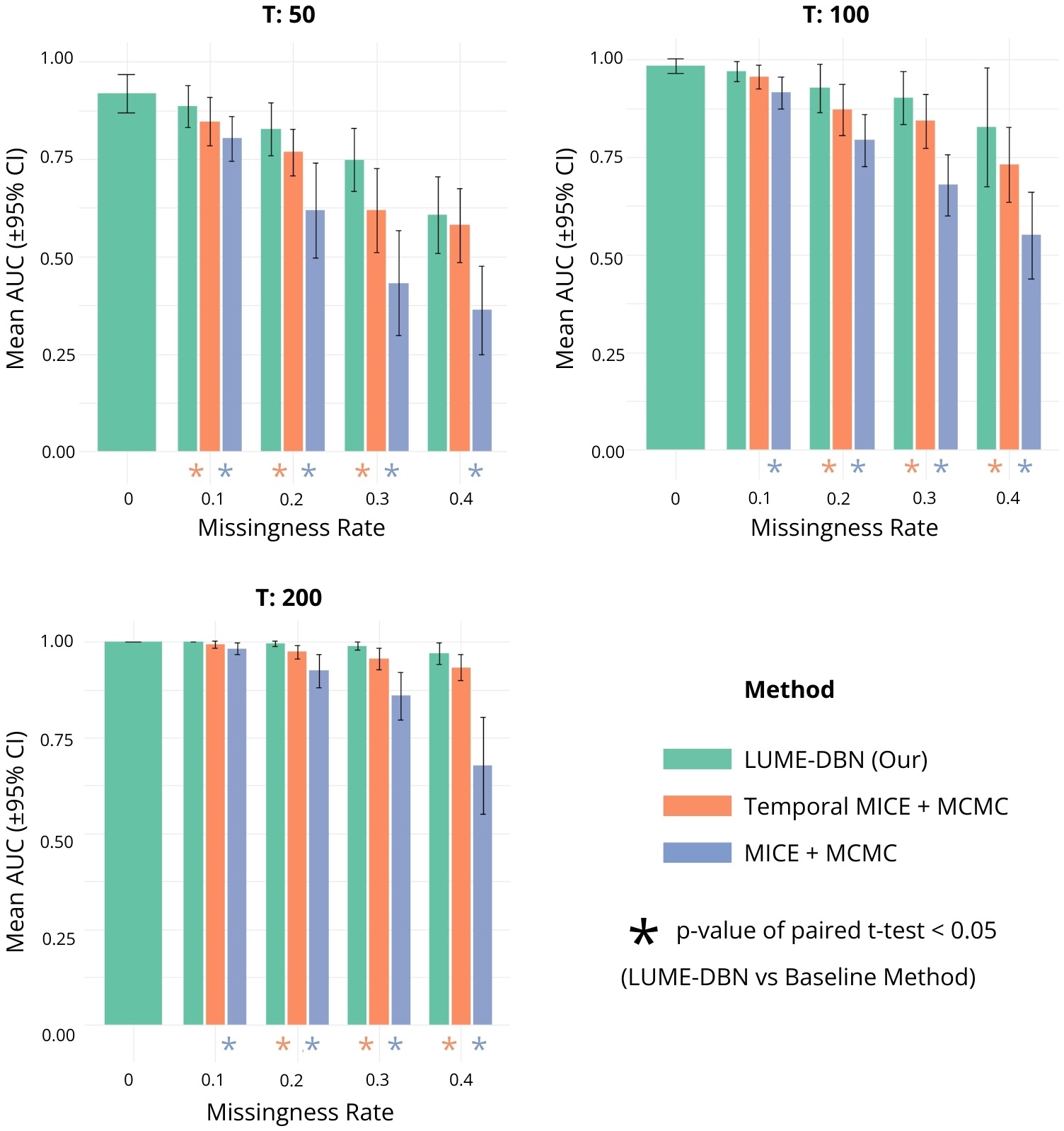}
    \caption{Area Under the Precision-Recall Curve for different experimental settings (sample sizes, missingness rates and imputation methods). The p-values of the paired t-test LUME-DBN AUC vs Baseline Method AUC are computed for each experimental condition, highlighting p-values < 0.05 with colored '$\star$' based on the baseline method. Confidence bars represent the 95\% confidence intervals for each experimental setting.}
    \label{fig:networkperformance}
\end{figure}

In Figure~\ref{fig:networkperformance} the results are shown aggregated for sample sizes and missingness rates. Confidence bars show the variability across different DBN models, whereas p-values of the paired t-test show the significance of the differences between each DBN's AUC-PR for LUME-DBN and for the baseline methods. The average AUC-PR on the complete data is reported to evaluate the impact of missingness in all the settings.

Across all settings, MICE consistently shows the worst performance, completely losing effectiveness once missingness exceeds $20\%$ reflecting its inefficiency when applied in the context of temporal data. Temporal MICE performs better, but remains significantly inferior to LUME-DBN across all sample sizes and missingness rates over $10\%$, as it is applied before the actual learning phase. In contrast, LUME-DBN proves more effective, particularly at high missingness levels, highlighting its superior ability to handle missing data, while not losing much compared to the fully observed case, especially for larger sample sizes.

\section{A case study in Intensive Care}
\label{sec:5}
\subsection{PhysioNet data description and preprocessing}
To validate our approach in the context of real-world intensive care data, we employ the PhysioNet 2012 Challenge dataset \cite{goldberger2012physionet}, a large-scale database comprising records from over 20,000 adult patients admitted to intensive care units (ICUs) across multiple hospitals. The dataset includes rich temporal and static information routinely collected in the ICU. All patients included in the dataset were hospitalized for at least 48 hours, and data collection covers the first 48 hours of ICU stay, with variables recorded at varying sampling frequencies. 

Since our approach is designed to manage only random missing values, knowing the systematic missingness due to patients' conditions we apply two consecutive preprocessing steps. First, we restrict our analysis on patients with severe clinical conditions, identified by a Sequential Organ Failure Assessment (SOFA) score greater than 12, consistent with thresholds used in clinical literature to define critical illness \cite{vincent1996sofa}. Secondly, we stratify patients into four subgroups \footnote{Medical ICU (MICU) $\rightarrow$ N = 34; Surgical ICU (SICU) $\rightarrow$ N = 104; Coronary Care Unit (CCU) $\rightarrow$ N = 114; Cardiac Surgery Recovery Unit (CSRU) $\rightarrow$ N = 62.} according to the ICU admission due to the diverse missingness rate per class. In such a way, assuming unconfoundness due to external reasons, we could consider missingness distribution as random conditional to the current model \cite{rubin1976inference}.

To handle the heterogeneity in measurement frequencies, we discretize the temporal data into 6-hour intervals, balancing high- and low-frequency signals following previous works on ICU monitoring and ensuring the tractability of missing value imputation based on the previous simulation results \cite{zhao2019temporal}. Following this, we apply a feature selection step based on missingness rates: variables with more than $40\%$ missing values in any ICU subgroup are excluded. This results in a final set of 11 clinical variables retained for analysis. The final set is comprised of vital signs, blood measurements, physiological features and a brain damage indicator, each described by temporal observations at $T = 9$ time points. 

Since variables differed in scale and variance, we standardized them using observed values before learning. Experiments are first runned with local standardization (per ICU group), then repeated with global standardization (entire dataset) to appreciate the impact of standardization on the entire learning procedure. Appendix~\ref{app:C} reports a description of the four ICU groups, the descriptive statistics of the included variables and the missingness rates by group.

\subsection{Convergence Diagnostics and Reconstructed DBNs}
Having demonstrated the superior reconstruction accuracy of LUME-DBN, we apply Algorithm~\ref{alg:learn-incomplete} to the ICU data. Appendix~\ref{app:D} reports the convergence of missing values and DBN structures, which is consistently reached within 5k epochs across all ICU groups. We then proceed under the same experimental conditions used in the simulations (missing values and parameter initialization, prior distributions, and burn-in restriction) for each ICU group, limiting the maximum number of parents to 5. Since no reference network is available for the real data, the validation of the method focuses on differences between the networks learned across groups and under different standardization strategies.

\begin{figure}[H]
    \centering
    \includegraphics[scale = 0.17]{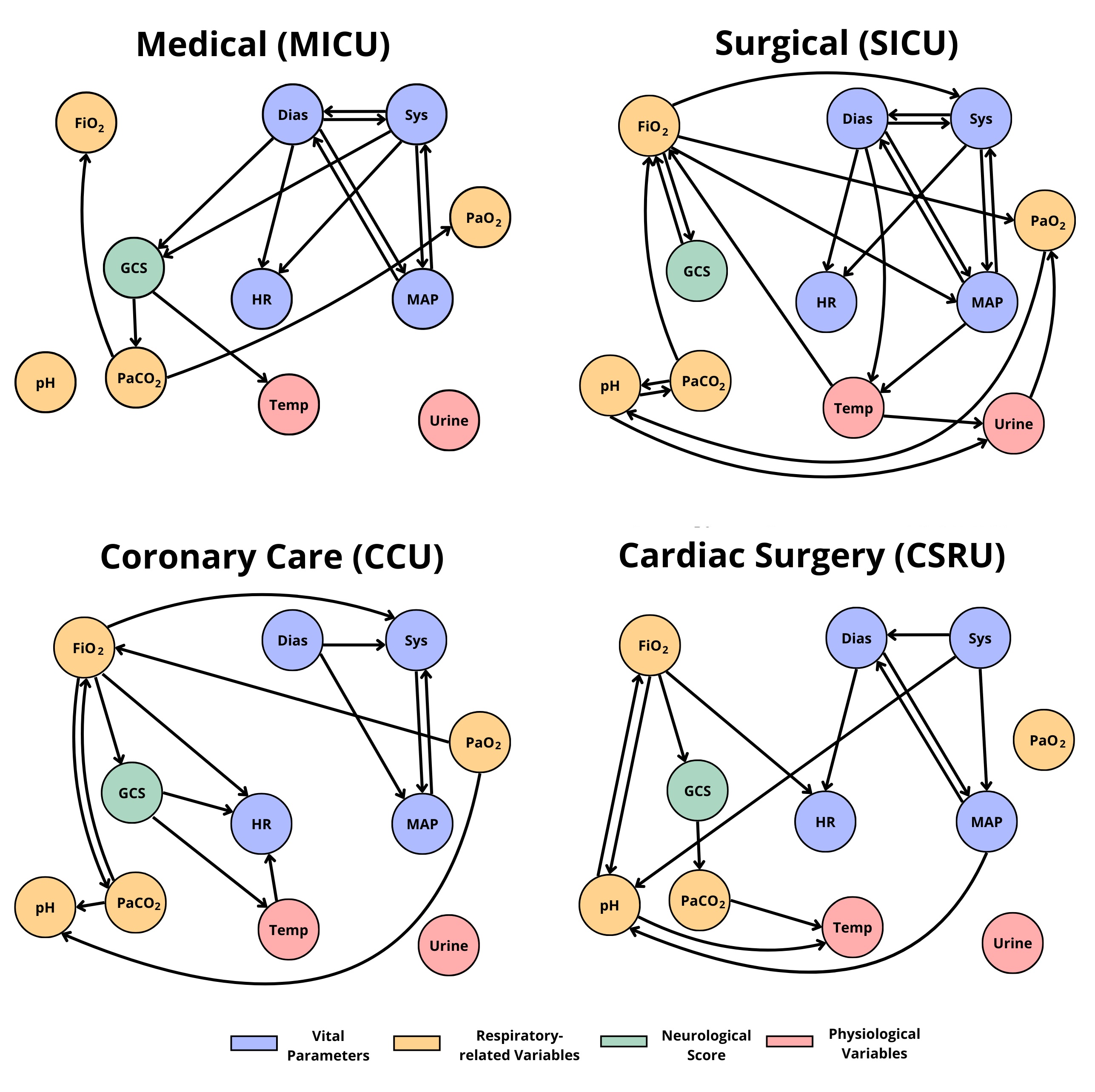}
    \caption{Reconstructed DBNs for each ICU type, averaged over five independent simulations after local data standardization. A threshold of 0.8 is applied to the averaged inclusion probabilities. Arcs are meant to represent temporal relationship with a single temporal lag, namely between nodes at time $t-1$ and nodes at time $t$.}
    \label{fig:ICULocalStandardization}
\end{figure}
 
In Figure \ref{fig:ICULocalStandardization} we report the reconstructed network structures obtained by applying a threshold of 0.8 to the inclusion probabilities averaged over the posterior samples to compare the reconstructed networks for the different groups. Supported by multiple surveys of critical care from the clinical literature \cite{hemodynamics, cardiovascularrespiratory}, we highlight the following commonalities and differences between different ICU groups in terms of temporal dynamics:

\begin{itemize}
    \item \textbf{Self-regulatory loops:}  
    Pressure parameters (MAP, Sys, Dias) were closely interrelated, reflecting the coordinated regulation of arterial blood pressure. Similarly, respiratory-related variables (FiO$_2$, PaCO$_2$, PaO$_2$, pH) were strongly connected, highlighting the integrated control of oxygenation, ventilation, and acid--base balance in all kind of ICUs.
    
    \item \textbf{Neurological interactions:}  
    Reduced consciousness increased heart rate in coronary care patients ($GCS \rightarrow HR$) and was linked to higher oxygen requirements in non-medical patients ($GCS \leftrightarrow FiO_2$), reflecting autonomic and respiratory coupling.
    
    \item \textbf{Hemodynamic effects:} Blood pressure strongly influenced consciousness ($Dias, MAP \rightarrow GCS$) in medical patients, underlining the importance of maintaining cerebral perfusion during shocks.
    
    \item \textbf{Thermoregulatory dynamics:} In surgical recovery, body temperature changes affected urine output ($Temp \rightarrow Urine$) and were also shaped by pressure and metabolic disturbances ($pH, PaCO_2 \rightarrow Temp$) after bypass. This highlights the sensitivity of thermoregulation to hemodynamic and metabolic conditions during post-operative care.
    
    \item \textbf{Cardiorespiratory feedbacks:} In coronary care and post-surgical patients, low oxygen levels triggered compensatory increases in heart rate ($FiO_2 \rightarrow HR$), consistent with hypoxia-driven cardiac responses.
\end{itemize}

Figure \ref{fig:reconstructednetworks} in Appendix~\ref{app:E} also reports results obtained with global standardization. In this case, the networks exhibit more commonalities, likely reflecting the uncertainty associated with small sample sizes and the interpretation of relationships as deviations from group-specific means. Certain temporal relationships appeared highly dependent on local standardization and are not well supported by clinical evidence, such as the influence of pressure parameters (Sys, MAP) on pH in CSRU patients or the effect of body temperature on $FiO_2$ levels in SICU patients. These findings highlight the need to incorporate prior clinical knowledge to guide and validate model inference.

\section{Conclusions and Research Directions}
\label{sec:6}
In this work we have proposed LUME-DBN, a novel method to learn DBNs from incomplete data positioned in a full Bayesian framework. We have derived the FCDs for missing values and showed they are tractable. Indeed we have employed them in the sampling-based imputation step to recover DBNs' structures. Our work highlights:
\begin{itemize}
    \item The convergence behaviour and effectiveness of LUME-DBN in learning DBN structures from both simulated and real-world ICU data, demonstrating superior reconstruction accuracy compared to model-agnostic methods.
    \item Significant improvements in terms of model interpretability due to its capacity of explicitly encoding uncertainty over the models, parameters and missing values' spaces.
    \item The potential for future extensions due to the Bayesian nature of the model, opening the possibility of incorporating prior clinical knowledge towards an expert-informed learning and improving the clinical relevance of the findings.
\end{itemize}

We also plan to compare our approach against classical frequentist approaches, but since DBN libraries do not yet support structure learning from incomplete data, dedicated development building on static BN libraries such as \textit{bnlearn}~\cite{bnlearn} will be required. An evaluation of the overall computation time required by the single steps of LUME-DBN will also be carried out, assessing how much time could be spared by parallelizing part of each module.

Another extension we have in mind is the integration of methods to manage Missing Not At Random (MNAR) patterns, which are frequent in clinical data due to unmodeled external factors. While methods combining BNs with missingness graphs have shown effectiveness in capturing MNAR mechanisms~\cite{liu2022greedy}, their applicability in temporal settings remain unexplored. Morover, extending these methods to continuous domains poses an additional challenge, as it requires dealing with mixed variable types. To address these complexities, we plan to investigate approaches based on logistic distributions, drawing inspiration from the global-optimization strategy of LiM~\cite{Zeng2022}.

We also aim to generalize our approach to non-homogeneous DBNs (NH-DBNs) to capture non-stationary relationships across time and patient groups. Specifically we plan to extend the method to globally coupled NH-DBNs, characterized by a shared structure among time and/or groups of patients, with deviations only in parameter distributions~\cite{shafiee2019nhgroups,zhang2022nhtime}. Such modeling is key to representing evolving treatment effects in intensive care.

\begin{credits}
\subsubsection{\ackname} 
This work is funded by the National National Plan for NRRP Complementary Investments (Project n. PNC0000003 - AdvaNced Technologies for Human-centrEd Medicine (ANTHEM)), and by the MUR under the grant ``Dipartimenti di Eccellenza 2023-2027'' of the Department of Informatics, Systems and Communication (DISCo), University of Milano-Bicocca, Milan, Italy.

\end{credits}

\printbibliography

\newpage
\appendix

\section*{Appendix} 
\addcontentsline{toc}{section}{Appendix} 
\setcounter{algorithm}{0}
\renewcommand{\thealgorithm}{\Alph{section}.\arabic{algorithm}}
\renewcommand{\thefigure}{\Alph{section}.\arabic{figure}}
\renewcommand{\thetable}{\Alph{section}.\arabic{table}}

\section{Full Conditional Distributions of Missing Data}
\label{app:A}
\begin{figure}[H]
    \centering
    \includegraphics[scale = 0.2]{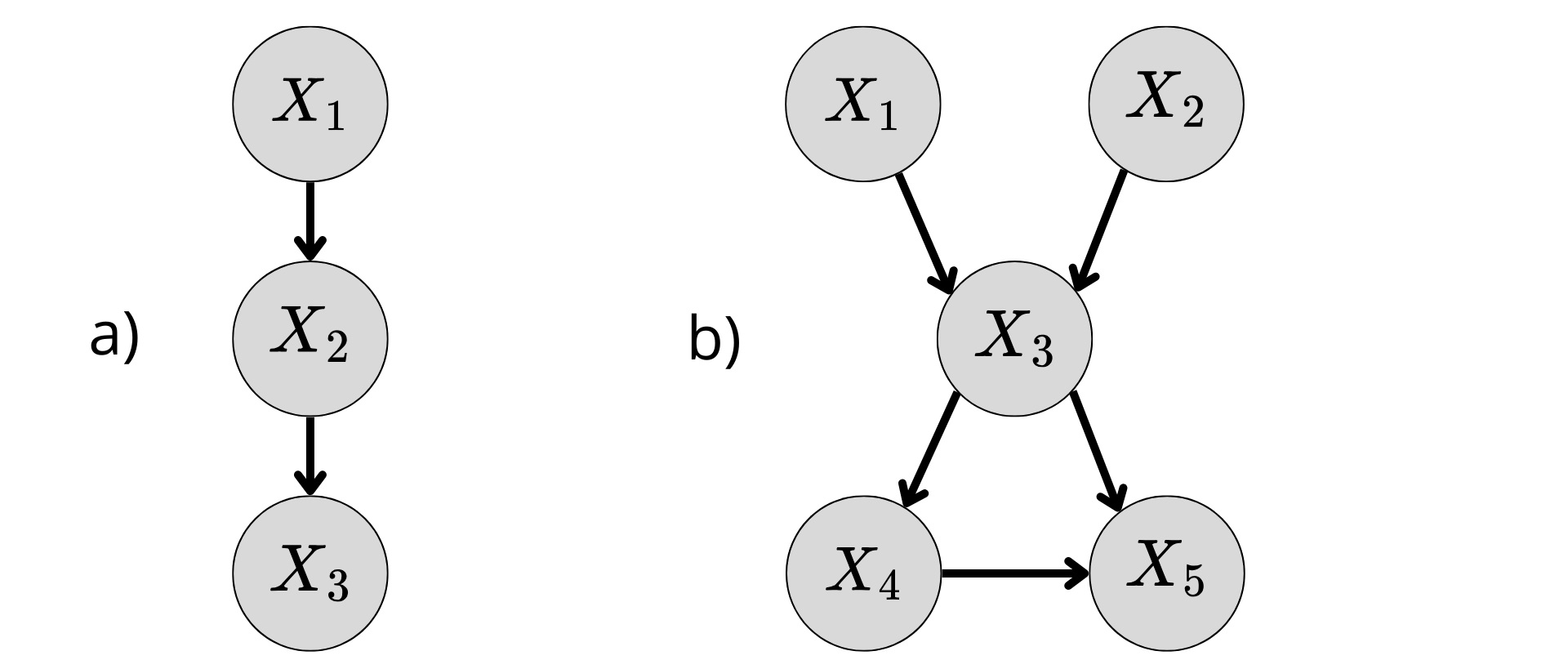}
    \caption{Two examples of DBNs. a) A DBN with 3 temporal nodes and 2 arcs. b) A more complex DBN with 5 temporal nodes, with a node $X_3^t$ with 2 parents $\{X_1^{t-1}, X_2^{t-1}\}$, 2 children $\{X_4^{t+1}, X_5^{t+1}\}$ and one node with a common children $\{X_4^{t}\}$ . }
    \label{fig:dbn_example}
\end{figure}

\noindent 
We consider three configurations where exactly one value is missing at time \( t = 2 \): \( {x_1^2}_{[MIS]} \), \({ x_2^2}_{[MIS]} \), or \({ x_3^2}_{[MIS]} \). The DBN is represented in Figure~\ref{fig:dbn_example}a and the depicted relationships are meant to represent temporal relationship with a single lag (from time $t-1$ to time $t$). These settings reflect structurally different roles in the DBN:  
\begin{itemize}
    \item \( X_1^2 \) has \textbf{no parents} and \textbf{one child},
    \item \( X_2^2 \) has \textbf{one parent and one child},
    \item \( X_3^2 \) has \textbf{one parent and no children}.
\end{itemize}

Assuming a Uniform prior over the domain of each variable, the FCD of a missing value only depends on the likelihood terms in which that variable appears. Since the samples are independent, missing values for different samples but for the same variable at the same time just lead to the same form of the Gaussian. \\

Consequently:
\begin{itemize}
    \item the FCD for \( X_3^2 \) depends only on its own likelihood: no demonstration is needed because the conditional likelihood of $X_3^2 |X_2^1$ is a properly defined Gaussian distribution with parameters ${\mu_3 = {\beta_0}^{(3)} + {\beta_1}^{(3)}X_2^1}$ and ${\sigma^2}^{(3)}$;
    \item the FCD for \( X_1^2 \) depends on its marginal likelihood and the conditional likelihood of its child \( X_2^3 \);
    \item the FCD for \( X_2^2 \) depends on both the conditional likelihood given its parent \( X_1^1 \) and the conditional likelihood of its child \( X_3^3 \).
\end{itemize}

\begin{align*}
P({x_1^2}_{[MIS]} \mid\,\cdot\,) 
&\propto P(X_1^2) \cdot P(X_2^3 \mid X_1^2) \\
&\propto \mathcal{N}({x_1^2}_{[MIS]} \mid \mu_1 = \beta_0^{(1)}, \sigma^{2(1)}) 
\cdot \mathcal{N}(x_2^3 \mid \mu_2 = \beta_0^{(2)}+\beta_1^{(2)} {x_1^2}_{[MIS]}, \sigma^{2(2)}) \\
&\propto \exp\left( - \frac{({x_1^2}_{[MIS]} - \beta_0^{(1)})^2}{2 \sigma^{2(1)}} \right) 
\cdot \exp\left( - \frac{(x_2^3 - \beta_0^{(2)} - \beta_1^{(2)} {x_1^2}_{[MIS]})^2}{2 \sigma^{2(2)}} \right) \\
&\propto \exp\left( - \frac{1}{2} \left[ 
\frac{({x_1^2}_{[MIS]})^2 - 2 {x_1^2}_{[MIS]} \beta_0^{(1)}}{\sigma^{2(1)}} 
+ \frac{(x_2^3)^2}{\sigma^{2(2)}} \right. \right. \\
&\quad \left. \left. 
- \frac{2 x_2^3 (\beta_0^{(2)} + \beta_1^{(2)} {x_1^2}_{[MIS]})}{\sigma^{2(2)}} 
+ \frac{(\beta_0^{(2)} + \beta_1^{(2)} {x_1^2}_{[MIS]})^2}{\sigma^{2(2)}}
\right] \right) \\
&\propto \exp\left( - \frac{1}{2} \left[
({x_1^2}_{[MIS]})^2 \left( \frac{1}{\sigma^{2(1)}} + \frac{(\beta_1^{(2)})^2}{\sigma^{2(2)}} \right) \right. \right.\\
&\quad \left. \left. + 2 {x_1^2}_{[MIS]} \left( \frac{-\beta_0^{(1)}}{\sigma^{2(1)}} + \frac{-\beta_1^{(2)}(x_2^3 - \beta_0^{(2)})}{\sigma^{2(2)}} \right) + C
\right] \right) \\
&\propto \mathcal{N} \left(
{x_1^2}_{[MIS]} \,\middle|\, 
\mu^* = {\sigma^2}^* \cdot \left(
\frac{\beta_0^{(1)}}{\sigma^{2(1)}} 
+ \frac{\beta_1^{(2)} (x_2^3 - \beta_0^{(2)})}{\sigma^{2(2)}} \right), \right. \\
&\quad \left. {\sigma^2}^* = 
\left(
\frac{1}{\sigma^{2(1)}} + \frac{(\beta_1^{(2)})^2}{\sigma^{2(2)}}
\right)^{-1}
\right)
\end{align*} 

\begin{align*}
P({x_2^2}_{[MIS]} \mid\,\cdot\,) 
&\propto P(X_2^2 \mid X_1^1) \cdot P(X_3^3 \mid X_2^2) \\
&\propto \mathcal{N}({x_2^2}_{[MIS]} \mid \mu_2 = \beta_0^{(2)}+\beta_1^{(2)} x_1^1, \sigma^{2(2)}) \\
& \quad \cdot \mathcal{N}(x_3^3 \mid \mu_3 = \beta_0^{(3)}+\beta_2^{(3)} {x_2^2}_{[MIS]}, \sigma^{2(3)}) \\ 
&\propto \exp\left( - \frac{({x_2^2}_{[MIS]} - \mu_2)^2}{2 \sigma^{2(2)}} \right) 
\cdot \exp\left( - \frac{(x_3^3 - \mu_3)^2}{2 \sigma^{2(3)}} \right) \\
&\propto \exp\left( -\frac{1}{2} \left[
({x_2^2}_{[MIS]})^2 \left( \frac{1}{\sigma^{2(2)}} + \frac{(\beta_2^{(3)})^2}{\sigma^{2(3)}} \right) \right. \right. \\
& \left. \left. + 2 {x_2^2}_{[MIS]} \left( \frac{-\beta_0^{(2)} - \beta_1^{(2)} x_1^1}{\sigma^{2(2)}} + \frac{-\beta_2^{(3)}(x_3^3 - \beta_0^{(3)})}{\sigma^{2(3)}} \right) + C
\right] \right) \\
&\propto \mathcal{N} \left(
{x_2^2}_{[MIS]} \,\middle|\,
\mu^* = {\sigma^2}^* \cdot
\left( \frac{\beta_0^{(2)} + \beta_1^{(2)} x_1^1}{\sigma^{2(2)}} 
+ \frac{\beta_2^{(3)} (x_3^3 - \beta_0^{(3)})}{\sigma^{2(3)}} \right), \right. \\
& \left. \quad{\sigma^2}^* = 
\left(
\frac{1}{\sigma^{2(2)}} + \frac{(\beta_2^{(3)})^2}{\sigma^{2(3)}}
\right)^{-1}
\right) 
\end{align*} \\

We then consider the DBN depicted in Figure~\ref{fig:dbn_example}b, and focus on a missing value \( {x_3^t}_{[MIS]} \) for variable \( X_3 \) at time \( t > 0 \). The Markov blanket of variable \( X_3^t \) is constituted of two parents (\( X_1^{t-1} \), \( X_2^{t-1} \)), two children (\( X_4^{t+1} \), \( X_5^{t+1} \)), and one variable (\( X_4^t \)) sharing a common child. Deriving the FCD for \( X_3^t \) in this setting illustrates the procedure in case of more complex Markov blankets, in the direction of demonstrating the tractability of the FCD in the broader case.

\begin{align*}
P({x_3^t}_{[MIS]} \mid\,\cdot\,) & \propto  P({X_3^t} \mid {X_1^{t-1}},  {X_2^{t-1}}) \,\cdot  P({X_4^{t+1}} \mid {X_3^{t}}) \,\cdot  P({X_5^{t+1}} \mid {X_3^{t}}, {X_4^{t}})\\
& \propto \mathcal{N}({x_3^t}_{[MIS]}|\mu_3= \beta_0^{(3)} + \beta_1^{(3)} x_1^{t-1} + \beta_2^{(3)} x_2^{t-1},{\sigma^2}^{(3)}) \\
& \cdot \mathcal{N}(x_4^{t+1}|\mu_4= \beta_0^{(4)} + \beta_3^{(4)} {x_3^t}_{[MIS]},{\sigma^2}^{(4)}) \\
& \cdot \mathcal{N}(x_5^{t+1}|\mu_5= \beta_0^{(5)} + \beta_3^{(5)} {x_3^t}_{[MIS]} + \beta_4^{(5)} x_4^t,{\sigma^2}^{(5)}) \\
& \propto \exp \left( - \frac{{(x_3^t}_{[MIS]} - (\beta_0^{(3)} + \beta_1^{(3)} x_1^{t-1} + \beta_2^{(3)} x_2^{t-1}))^2}{2 {\sigma^2}^{(3)}}\right) \\
& \cdot \exp \left( - \frac{(x_4^{t+1} - (\beta_0^{(4)} + \beta_3^{(4)} {x_3^t}_{[MIS]}))^2}{2 {\sigma^2}^{(4)}}\right) \\
& \cdot \exp \left( - \frac{(x_5^{t+1} - (\beta_0^{(5)} + \beta_3^{(5)} {x_3^t}_{[MIS]} + \beta_4^{(5)} x_4^t))^2}{2 {\sigma^2}^{(5)}}\right) 
\end{align*} \\

\noindent
We define $\mu_{\{-3\}}^{(4)} := \mu^{(4)} - \beta_3^{(4)} {x_3^t}_{[MIS]}$ and $\mu_{\{-3\}}^{(5)} := \mu^{(5)} - \beta_3^{(5)} {x_3^t}_{[MIS]}$. Then $(x_4^{t+1} - (\beta_0^{(4)} + \beta_3^{(4)} {x_3^t}_{[MIS]}))^2 = (x_4^{t+1} - \beta_3^{(4)} {x_3^t}_{[MIS]} - \mu_{\{-3\}}^{(4)})^2 $ and $(x_5^{t+1} - (\beta_0^{(5)} + \beta_3^{(5)} {x_3^t}_{[MIS]} + \beta_4^{(5)} x_4^t))^2 = (x_5^{t+1} - \beta_3^{(5)} {x_3^t}_{[MIS]} - \mu_{\{-3\}}^{(5)})^2$. For \( j \in \{4, 5\} \), each term \( (x_j^{t+1} - \beta_3^{(j)} {x_3^t}_{[\text{MIS}]} - \mu_{\{-3\}}^{(j)})^2 \) is a squared trinomial. Since we are computing the FCD for \( {x_3^t}_{[\text{MIS}]} \), we can focus on the terms that involve this quantity.
Namely:
\[
(x_j^{t+1} - \beta_3^{(j)} {x_3^t}_{[\text{MIS}]} - \mu_{\{-3\}}^{(j)})^2 \propto
({x_3^t}_{[\text{MIS}]}\beta_3^{(j)})^2 
- 2 \beta_3^{(j)} {x_3^t}_{[\text{MIS}]} (x_j^{t+1} - \mu_{\{-3\}}^{(j)}) 
\] \\
Indeed, we could rewrite $({x_3^t}_{[MIS]} - (\beta_0^{(3)} + \beta_1^{(3)} x_1^{t-1} + \beta_2^{(3)} x_2^{t-1}))^2 = {(x_3^t}_{[MIS]} - \mu_3)^2$. Again, we are interested in the terms of the squared binomial involving ${x_3^t}_{[MIS]}$, thus: \\

\[
({x_3^t}_{[MIS]} - \mu_3)^2 \propto {({x_3^t}_{[MIS]})}^2 - 2 {x_3^t}_{[MIS]} \mu_3
\] \\
\noindent
Then: \\

\begin{align*}
P({x_3^t}_{[MIS]} \mid\,\cdot\,) & \propto \exp \left( - \frac{1}{2} \left( ({x_3^t}_{[MIS]})^2 \cdot \left(\frac{1}{{\sigma^2}^{(3)}}\right) - (2{x_3^t}_{[MIS]}) \left( \frac{\mu_3}{{\sigma^2}^{(3)}} \right)\right) \right) \\
& \cdot \exp \left( - \frac{1}{2} \left( ({x_3^t}_{[MIS]})^2 \cdot \left(\frac{(\beta_3^{(4)})^2}{{\sigma^2}^{(4)}}\right) - (2{x_3^t}_{[MIS]}) \left( \frac{\beta_3^{(4)} (x_4^{t+1} - \mu_{\{-3\}}^{(4)})}{{\sigma^2}^{(4)}} \right)\right) \right) \\
& \cdot \exp \left( - \frac{1}{2} \left( ({x_3^t}_{[MIS]})^2 \cdot \left(\frac{(\beta_3^{(5)})^2}{{\sigma^2}^{(5)}}\right) - (2{x_3^t}_{[MIS]}) \left( \frac{\beta_3^{(5)} (x_5^{t+1} - \mu_{\{-3\}}^{(5)})}{{\sigma^2}^{(5)}} \right)\right) \right) \\
& \propto \exp \left( - \frac{1}{2} \left( ({x_3^t}_{[MIS]})^2 \cdot \left(\frac{1}{{\sigma^2}^{(3)}} + \frac{(\beta_3^{(4)})^2}{{\sigma^2}^{(4)}} + \frac{(\beta_3^{(5)})^2}{{\sigma^2}^{(5)}} \right) + \right. \right. \\
& \left. \left. - (2{x_3^t}_{[MIS]}) \left( \frac{\mu_3}{{\sigma^2}^{(3)}} + \frac{\beta_3^{(4)} (x_4^{t+1} - \mu_{\{-3\}}^{(4)})}{{\sigma^2}^{(4)}} + \frac{\beta_3^{(5)} (x_5^{t+1} - \mu_{\{-3\}}^{(5)})}{{\sigma^2}^{(5)}}\right) + C \right) \right) \\ 
& \propto \mathcal{N}\left(\mu^*= {\sigma^2}^* \cdot \left( \frac{\mu_3}{{\sigma^2}^{(3)}} + \frac{\beta_3^{(4)} (x_4^{t+1} - \mu_{\{-3\}}^{(4)})}{{\sigma^2}^{(4)}} + \frac{\beta_3^{(5)} (x_5^{t+1} - \mu_{\{-3\}}^{(5)})}{{\sigma^2}^{(5)}}\right), \right. \\ 
& \left. {\sigma^2}^*=\left(\frac{1}{{\sigma^2}^{(3)}} + \frac{(\beta_3^{(4)})^2}{{\sigma^2}^{(4)}} + \frac{(\beta_3^{(5)})^2}{{\sigma^2}^{(5)}} \right)^{-1} \right) \\
\end{align*} \\

We first derived the FCD for a missing value at a specific time in a simple DBN with three nodes. We then illustrated its form for a variable with two parents and two children at a generic time $t$. Since all FCDs are Gaussian distributions with closed-form parameters, we now present the general case for a missing value on a variable $X_i$ at time $t$, characterized by $n$ parents $\pi_{(i)}=\{X^{t-1}_{p_1}, \dots,X^{t-1}_{p_n}\}$ and $m$ children $\{X^{t+1}_{c_1}, \dots,X^{t+1}_{c_n}\}$..

The FCD's of the missing value ${x^t_i}_{[MIS]}$ depends on the conditional likelihood of $X_i$ and the conditional likelihoods of its children. As before, we isolate the contribution of ${x^t_i}_{[MIS]}$ in its children likelihoods. Specifically, for each child $j \in \{X^{t+1}_{c_1}, \dots,X^{t+1}_{c_n}\}$, given the mean term $\mu^{t+1}_j$ and the linear coefficient $\beta^{(j)}_i$ associated with the Gaussian likelihood of $X_j$, we define 

\[
{\mu^{(j)}_{\{-i\}}}^{(t+1)} = \mu^{t+1}_j \,-\,\beta^{(j)}_i {x_i^t}_{[MIS]}
\]
\\
We can then derive the missing value FCD as follows:

\begin{align*}
P({x_i^t}_{[MIS]} \mid\,\cdot\,) & \propto  P(X_i^t \,| \, \pi_{(i)}) \,\cdot\,\prod_{j \in \{c_1,\dots,c_m\}} P(X^{t+1}_j|\pi_{(j)}) \\
&\propto \mathcal{N} ({x^t_i}_{[MIS]} |\mu^t_{i} = \beta^{(i)}_0 + \beta^{(i)}_{1} x^{t-1}_{p_1} + \dots + \beta^{(i)}_{n} x^{t-1}_{p_n}, {\sigma^2}^{(i)}) \\
&\cdot\, \prod_{j \in \{c_1,\dots,c_m\}} \mathcal{N} ({x^{t+1}_j} |\mu^{t+1}_{j} = {\mu^{(j)}_{\{-i\}}}^{(t+1)} + \beta^{(j)}_i {x_i^t}_{[MIS]}, {\sigma^2}^{(j)}) \\
& \propto \exp \left( - \frac{1}{2} \left( ({x_i^t}_{[MIS]})^2 \cdot \left(\frac{1}{{\sigma^2}^{(i)}}\right) - (2{x_i^t}_{[MIS]}) \left( \frac{\mu^t_i}{{\sigma^2}^{(i)}} \right)\right) \right) \\
& \cdot\,  \prod_{j \in \{c_1,\dots,c_m\}}\exp \left( - \frac{1}{2} \left( ({x_i^t}_{[MIS]})^2 \cdot \left(\frac{(\beta_i^{(j)})^2}{{\sigma^2}^{(j)}}\right) \right. \right. + \\ 
& \left. \left. - (2{x_i^t}_{[MIS]}) \left( \frac{\beta_i^{(j)} (x_j^{t+1} - {\mu_{\{-i\}}^{(j)}}^{(t+1)})}{{\sigma^2}^{(j)}} \right)\right) \right) \\
& \propto \exp \left( - \frac{1}{2} \left( ({x_i^t}_{[MIS]})^2 \cdot \left(\frac{1}{{\sigma^2}^{(i)}} + \sum_{j \in \{c_1,\dots,c_m\}} \frac{(\beta_i^{(j)})^2}{{\sigma^2}^{(j)}}  \right) + \right. \right. \\
& \left. \left. - (2{x_i^t}_{[MIS]}) \left( \frac{\mu^t_i}{{\sigma^2}^{(i)}} +  \sum_{j \in \{c_1,\dots,c_m\}} \frac{\beta_i^{(j)} (x_j^{(t+1)} - {\mu_{\{-i\}}^{(j)}}^{(t+1)})}{{\sigma^2}^{(j)}}  \right) + C \right) \right) \\ 
& \propto \mathcal{N}\left(\mu^*= {\sigma^2}^* \cdot \left( \frac{\mu^t_i}{{\sigma^2}^{(i)}} +  \sum_{j \in \{c_1,\dots,c_m\}} \beta_i^{(j)} \frac{(x_j^{t+1}- {\mu^{(j)}_{\{-i\}}}^{(t+1)})}{{\sigma^2}^{(j)}} \right), \right. \\ 
& \left. {\sigma^2}^*=\left( \frac{1}{{\sigma^2}^{(i)}} + \sum_{j \in \{c_1,\dots,c_m\}} \frac{{(\beta_i^{(j)})}^2}{{\sigma^2}^{(j)}} \right)^{-1} \right) 
\end{align*}

We can now express the FCD of a generic missing value ${x_i^t}_{[MIS]}$ for a variable $X_i$ at time $t$ in the context of a DBN, accounting for an arbitrary number of parents and children:

\begin{align*}
P({x_i^t}_{[MIS]} \mid \cdot)
&= \mathcal{N}\bigl(\mu^*, {\sigma^2}^*\bigr) \label{eq:conditional_missing}\\
\text{where:}\quad
&\begin{aligned}[t]
&{\sigma^2}^* = \left(
\frac{1}{{\sigma^2}^{(i)}} + \displaystyle\sum_{j:\; (X_i^t \in \pi_{(j)})}
\frac{(\beta_i^{(j)})^2}{{\sigma^2}^{(j)}}
\right)^{-1}, \\
&\mu^* = {\sigma^2}^* \cdot \left(
\frac{\mu^t_i}{{\sigma^2}^{(i)}} + \displaystyle\sum_{j:\; (X_i^t \in \pi_{(j)})}
\beta_i^{(j)} \frac{\bigl(x_j^{t+1}- {\mu^{(j)}_{\{-i\}}}^{(t+1)}\bigr)}{{\sigma^2}^{(j)}}
\right).
\end{aligned}
\end{align*}

\section{Learning DBNs from Incomplete Data}
\label{app:B}

\begin{algorithm}[H]
\caption{Parameter Set Update via Collapsed Gibbs Sampling}
\label{alg:learn-params}
\resizebox{\linewidth}{!}{%
\begin{minipage}{\linewidth}
\setstretch{1.3} 
\begin{algorithmic}[1]
\State \textbf{Input:} Data $\{Y, X\}$, priors $\{\alpha_\sigma, \beta_\sigma, a, b, \mu\}$, current $\{\sigma^2,\delta^2,\pi\}$
\State \textbf{Sample $\sigma^{2}$ (Collapsed Gibbs step)}
\setlength{\abovedisplayskip}{3pt}
\setlength{\belowdisplayskip}{3pt}
\[
\sigma^2 \sim \mathrm{Inv\!-\!GAM} \Big( 
  \alpha_\sigma + \frac{N T}{2},\ 
  \beta_\sigma + \frac12 (Y - X_{[\pi]}\mu_{[\pi]})^\top 
  ( I + \sigma^2 X_{[\pi]} X_{[\pi]}^\top )^{-1}
  (Y - X_{[\pi]}\mu_{[\pi]})
\Big)
\]
\State \textbf{Sample $\beta$}
\[
\beta \sim \mathcal{N} \Big(
  (\sigma^{-2} I + X_{[\pi]}^\top X_{[\pi]})^{-1}
  (\sigma^{-2} \mu_{[\pi]} + X_{[\pi]}^\top Y),\ 
  \sigma^2 (\sigma^{-2} I + X_{[\pi]}^\top X_{[\pi]})^{-1}
\Big)
\]
\State \textbf{Sample $\delta^{2}$}
\[
\delta^{2} \sim \mathrm{Inv\!-\!GAM} \Big( 
  a + \frac{|\pi| + 1}{2},\ 
  b + \frac12 \sigma^{-2} 
  (\beta - \mu_{[\pi]})^\top (\beta - \mu_{[\pi]})
\Big)
\]
\State \textbf{Output:} Posterior sample $\{\beta, \sigma^2, \delta^2\}$
\end{algorithmic}
\end{minipage}%
}
\end{algorithm}

\begin{algorithm}[H]
\caption{Covariate Set Update via Metropolis--Hastings}
\label{alg:learn-parents}
\resizebox{\linewidth}{!}{%
\begin{minipage}{\linewidth}
\setstretch{1.3} 
\begin{algorithmic}[1]
\State \textbf{Input:} Data $\{Y, X\}$, current covariate set $\pi$, parameters $\{\delta^2, \sigma^2\}$
\State Choose a move type (Deletion, Addition, or Exchange) uniformly at random
\State Generate candidate covariate set $\pi_{\star}$ based on the chosen move
\State Compute acceptance probability:
\[
A(\pi \rightarrow \pi_{\star}) = \min \left\{1,\ 
\frac{p(Y \mid \pi_{\star}, \delta^2)}{p(Y \mid \pi, \delta^2)} \cdot 
\frac{p(\pi_{\star})}{p(\pi)} \cdot 
HR
\right\}
\]
with $HR = \tfrac{|\pi|}{n-|\pi_{\star}|}$ (Deletion), 
$\tfrac{n-|\pi|}{|\pi_{\star}|}$ (Addition), or $1$ (Exchange); \qquad \qquad 
$p(Y \mid \delta^2, \pi) =
\frac{\Gamma\left( A \right)}{\Gamma(\alpha_\sigma)} \cdot
\pi^{-\frac{T}{2}} (2\beta_\sigma)^{\alpha_\sigma}
\det(C)^{-1/2}
\left( 2\beta_\sigma + M^\top
C^{-1} M \right)^{-A}
$. \footnote{$A = \frac{T}{2}+\alpha_\sigma$, $C =I +\delta^2 XX^T$, $M = Y - X\mu$}
\State Draw $u \sim \mathcal{U}(0, 1)$
\If{$u < A$}
    \State Accept: $\pi \gets \pi_{\star}$
    \setlength{\abovedisplayskip}{3pt}
    \setlength{\belowdisplayskip}{3pt}
    \State Sample $\beta$ from FCD:
    \[
    \beta \sim \mathcal{N} \Big(
      (\sigma^{-2} I + X_{[\pi]}^\top X_{[\pi]})^{-1}
      (\sigma^{-2} \mu_{[\pi]} + X_{[\pi]}^\top Y),\ 
      \sigma^2 (\sigma^{-2} I + X_{[\pi]}^\top X_{[\pi]})^{-1}
    \Big)
    \]
\EndIf
\State \textbf{Output:} Posterior sample $\{\pi, \beta\}$
\end{algorithmic}
\end{minipage}%
}
\end{algorithm}

\begin{algorithm}[H]
\caption{Missing Data Update via Gibbs Sampling}
\label{alg:learn-missings}
\begin{spacing}{1.4}
\begin{algorithmic}[1]
\State \textbf{Input:} Missing Variable Index $i$, \\ 
\quad \quad \quad \quad $\mathcal{D} = [x_j^t]_{j =1:k, t = 1:3} $, $\mathcal{B} = [\beta_i^{(j)}]_{i =0:k, j = 1:k}$, $\overline{\Sigma} = \{{\sigma^2}^{(j)}\}_{j =1:k}$
\State Compute prior mean:
\[
\mu_i= \sum_{p: (X_p^{1} \in  \pi_{(i)})} x_p^{1} \beta^{(i)}_p
\]
\State For each $j$, compute:
\[
{\mu^{(j)}_{\{-i\}}} = \sum_{m \ne i: (X_m^2 \in \pi_{(j)}) } x_m^2 \cdot \beta_m^{(j)}
\]
\State Compute posterior variance:
\[
{\sigma^2}^* = \left( \frac{1}{{\sigma^2}^{(i)}} + \sum_{j: (X_i^2 \in \pi_{(j)})} \frac{{(\beta_i^{(j)})}^2}{{\sigma^2}^{(j)}} \right)^{-1}
\]
\State Compute posterior mean:
\[
\mu^* = {\sigma^2}^* \cdot \left( \frac{\mu_i}{{\sigma^2}^{(i)}} + \sum_{j: (X_i^2 \in \pi_{(j)})} \beta_i^{(j)} \cdot \frac{x_j^{3} - {\mu^{(j)}_{\{-i\}}}}{{\sigma^2}^{(j)}} \right)
\]
\State Sample ${{x_i}_{[MIS]}} \sim \mathcal{N}(\mu^*, {\sigma^2}^*)$
\State Output: Posterior Sample ${{x_i}_{[MIS]}}$
\end{algorithmic}
\end{spacing}
\end{algorithm}

\begin{algorithm}[H]
\caption{LUME-DBN}
\label{alg:learn-incomplete}
\begin{spacing}{1.4}
\begin{algorithmic}[1]
\State \textbf{Input:} Incomplete Dataset $\mathcal{D}_M = [{x^t_{i,j}}]_{i = 1:N; j = 1:k; t = 1:T}$, number of epochs $E$, Missing Imputation freq $E_{M}$, prior parameters {$\{\alpha_\sigma, \beta_\sigma, a, b, \mu = \{\mu^{(j)}_i\}_{i=0:k,j=1:k} \}$}
\State \textbf{Initialize:} Initial models $\mathcal{M}_{(0)} = [{\pi_i^{(j)}}_{(0)}]_{i, j = 1:k}$, \\ \quad \quad \quad \quad \quad \ Initial linear coefficients $\mathcal{B}_{(0)} = [{\beta_i^{(j)}}_{(0)}]_{i = 0:k; j = 1:k}$, \\ \quad \quad \quad \quad \quad \ Initial noise parameters $\overline{\Sigma}_{(0)} = \{{\sigma^2}_{(0)}^{(j)}\}_{j = 1:k}$, \\ \quad \quad \quad \quad \quad \ Initial uncertainty parameters $\overline{\Delta}_{(0)} = \{{\delta^2}_{(0)}^{(j)}\}_{j = 1:k}$, \\ \quad \quad \quad \quad \quad \ Initial completed dataset $\mathcal{D}_{(0)} = [{x^t_{i,j}}_{[MIS]}^{(0)}]_{i = 1:N; j = 1:k; t = 1:T}$\\
$e \gets 0$
\While{e $<$ E}
    \State $e \gets e + 1$
    \State $\mathcal{X}_{(e)} \gets Lag(\mathcal{D}_{(e-1)})$
    \For{$e_m  \text{ in } \{1, \dots,E_M\}$} 
        \State $e \gets e + 1$
        \For{$j  \text{ in } \{1, \dots,k\}$} 
            \State $Y \gets X^t_j$; \quad $X \gets \mathcal{X}^{(t-1)}_{\{-j\}}$; \quad $\mu \gets \mu^{(j)}$;
            \State $\pi \gets {\pi^{(j)}}_{(e-1)}$, \quad $\beta \gets \beta^{(j)}_{(e-1)}$; \quad $\sigma^2 \gets {\sigma^2}^{(j)}_{(e-1)}$; \quad $\delta^2 \gets{\delta^2}^{(j)}_{(e-1)}$
            \State {\textbf{Parameters Move}: }
                \State Update via \textbf{Algorithm B.1}: $\{{\beta^{(j)}}_{(e)}, {\sigma^2}_{(e)}^{(j)},{\delta^2}_{(e)}^{(j)} \} \gets \{\beta, \sigma^2,\delta^2 \}$
            \State {\textbf{Model Move}: }
                \State Update via \textbf{Algorithm B.2}: $\{{\pi^{(j)}}_{(e)}, \beta^{(j)}_{(e)} \} \gets \{\pi, \beta\}$
        \EndFor
    \EndFor
    \State $\mathcal{M}_{(e)} \gets [{\pi_i^{(j)}}_{(e)}]_{i, j = 1:k}$,  \quad $\mathcal{B}_{(e)} \gets [{\beta_i^{(j)}}_{(e)}]_{i = 0:k; j = 1:k}$
    \State $\overline{\Sigma}_{(e)} \gets \{{\sigma^2}_{(e)}^{(j)}\}_{j = 1:k}$, \quad $\overline{\Delta}_{(e)} \gets \{{\delta^2}_{(e)}^{(j)}\}_{j = 1:k}$
    \State {\textbf{Missing Data Imputation Move}:}
    \For{$\{t,i\}: \exists {x^t_{i,j}} = \text{NA}$  (with $j \in \{1,\dots,k\}$} 
    \State $\mathcal{D} \gets [{x^t_{i,j}}_{[MIS]}^{(e-1)}]_{j = 1:k; t = (t-1:t+1)}$
    \State $\mathcal{M} \gets \mathcal{M}_{(e)} $, \quad $\mathcal{B} \gets \mathcal{B}_{(e)} $,  \quad $\overline{\Sigma} \gets \overline{\Sigma}_{(e)}$
        \For{$j: {x^t_{i,j}} $ is missing} 
        \State Update Missing via \textbf{Algorithm B.3}: ${x^t_{i,j}}_{[MIS]}^{(e)} \gets {x_{j}}_{[MIS]}$
        \EndFor
    \EndFor
    \State $\mathcal{D}_{(e)} \gets [{x^t_{i,j}}_{[MIS]}^{(e)}]_{i = 1:N; j = 1:k; t = 1:T}$
\EndWhile
\State \textbf{Output:} Posterior samples $\{\mathcal{M}_{(e)}, \mathcal{B}_{(e)}, \overline{\Sigma}_{(e)}, \overline{\Delta}_{(e)},\mathcal{D}_{(e)}\}_{e=0:E}$
\end{algorithmic}
\end{spacing}
\end{algorithm}

\section{Intensive Care Data Description}
\label{app:C}
\begin{table}[H]
\centering
\setlength{\tabcolsep}{3pt} 
\renewcommand{\arraystretch}{2} 
\caption{Descriptions of ICU subgroups included in the study along with the number of patients per group.}
\label{tab:icu_subgroups}
\begin{tabular}{@{} l p{8cm} c @{}}
\toprule
\textbf{Group} & \textbf{Description} & \textbf{Number of Patients} \\
\midrule
MICU & Medical ICU: Patients having medical conditions requiring intensive monitoring and treatment (i.e. Sepsis, Pneumonia, ...). & 34 \\
\addlinespace
SICU & Surgical ICU: Post-operative patients recovering from non-cardiac surgeries (abdominal, vascular, orthopedic, or general surgeries). Monitored to avoid bleeding, infection, or organ dysfunction post-surgery. & 104 \\
\addlinespace
CCU & Coronary Care Unit: Patients in intensive care for acute cardiac events and conditions like Myocardial infarction, Arrhythmias or Acute heart failure, which do not need surgeries. & 114 \\
\addlinespace
CSRU & Cardiac Surgery Recovery Unit: Patients recovering from cardiac surgeries (i.e., Coronary artery bypass grafting, valve replacements, aortic repairs). Under observation to ensure hemodynamic stability and recovery after major heart operations. & 62 \\
\bottomrule
\addlinespace
\addlinespace
\end{tabular}
\end{table}

\begin{table}[H]
\centering
\setlength{\tabcolsep}{9pt} 
\renewcommand{\arraystretch}{2} 
\caption{Clinical variables and their missingness rates (\%) across ICU subgroups. Variables with more than 40\% missingness in any group have been excluded.}
\label{tab:missingness_rates}
\begin{tabular}{@{} l p{4.5cm} l l l l @{}}
\toprule
\textbf{Variable} & \textbf{Description} & \textbf{MICU} & \textbf{SICU} & \textbf{CCU} & \textbf{CSRU} \\
\midrule
HR & Heart Rate (beats per minute) & 0.98 & 6.52 & 0.39 & 1.79 \\
\addlinespace
Dias & Diastolic Arterial Blood Pressure (mmHg) & 6.54 & 10.15 & 20.57 & 3.94 \\
\addlinespace
MAP & Mean Arterial Pressure (mmHg) & 6.54 & 10.04 & 21.44 & 3.94 \\
\addlinespace
Sys & Systolic Arterial Blood Pressure (mmHg) & 6.54 & 10.15 & 20.57 & 3.94 \\
\addlinespace
Temp & Core Body Temperature (°C) & 9.48 & 8.01 & 4.78 & 3.76 \\
\addlinespace
$FiO_2$ & Fraction of Inspired Oxygen (\%) & 10.46 & 33.01 & 13.06 & 13.08 \\
\addlinespace
GCS & Glasgow Coma Scale (3–15), neurological assessment score & 14.05 & 14.64 & 5.85 & 6.63 \\
\addlinespace
Urine & Hourly Urine Output (mL), measure of renal function & 19.28 & 11.54 & 7.21 & 6.81 \\
\addlinespace
pH & Arterial Blood pH, indicator of acid–base status & 20.59 & 15.92 & 35.28 & 16.67 \\
\addlinespace
$PaCO_2$ & Partial Pressure of Carbon Dioxide in Arterial Blood (mmHg) & 20.91 & 19.44 & 35.67 & 16.67 \\
\addlinespace
$PaO_2$ & Partial Pressure of Oxygen in Arterial Blood (mmHg) & 21.24 & 19.34 & 35.67 & 16.67 \\
\bottomrule
\addlinespace
\addlinespace
\end{tabular}
\end{table}

\begin{table}[H]
\centering
\setlength{\tabcolsep}{12pt} 
\renewcommand{\arraystretch}{3} 
\caption{Mean ($\mu$) and variance ($\sigma^2$) of clinical variables across ICU subgroups.}
\label{meansandvariances}
\begin{tabular}{@{} l l l l l l @{}}
\toprule
\textbf{Variable} & \textbf{MICU} & \textbf{SICU} & \textbf{CCU} & \textbf{CSRU} & \textbf{All} \\
\midrule
HR & \makecell[l]{$\mu$: 91 \\ $\sigma^2$: 491} & \makecell[l]{$\mu$: 87 \\ $\sigma^2$: 170} & \makecell[l]{$\mu$: 91 \\ $\sigma^2$: 336} & \makecell[l]{$\mu$: 93 \\ $\sigma^2$: 409} & \makecell[l]{$\mu$: 91 \\ $\sigma^2$: 332} \\
Dias & \makecell[l]{$\mu$: 55.60 \\ $\sigma^2$: 131.60} & \makecell[l]{$\mu$: 55.74 \\ $\sigma^2$: 75.94} & \makecell[l]{$\mu$: 60.83 \\ $\sigma^2$: 128.98} & \makecell[l]{$\mu$: 58.85 \\ $\sigma^2$: 160.97} & \makecell[l]{$\mu$: 57.84 \\ $\sigma^2$: 126.01} \\
MAP & \makecell[l]{$\mu$: 74.69 \\ $\sigma^2$: 129.16} & \makecell[l]{$\mu$: 74.07 \\ $\sigma^2$: 82.05} & \makecell[l]{$\mu$: 80.87 \\ $\sigma^2$: 184.04} & \makecell[l]{$\mu$: 78.56 \\ $\sigma^2$: 410} & \makecell[l]{$\mu$: 77.08 \\ $\sigma^2$: 224.13} \\
Sys & \makecell[l]{$\mu$: 104 \\ $\sigma^2$: 397} & \makecell[l]{$\mu$: 113 \\ $\sigma^2$: 192} & \makecell[l]{$\mu$: 121 \\ $\sigma^2$: 445} & \makecell[l]{$\mu$: 108 \\ $\sigma^2$: 460} & \makecell[l]{$\mu$: 112 \\ $\sigma^2$: 389} \\
Temp & \makecell[l]{$\mu$: 37.02 \\ $\sigma^2$: 6.66} & \makecell[l]{$\mu$: 37.08 \\ $\sigma^2$: 1.25} & \makecell[l]{$\mu$: 36.79 \\ $\sigma^2$: 2.87} & \makecell[l]{$\mu$: 36.86 \\ $\sigma^2$: 0.95} & \makecell[l]{$\mu$: 36.93 \\ $\sigma^2$: 2.04} \\
FiO2 & \makecell[l]{$\mu$: 0.57 \\ $\sigma^2$: 0.03} & \makecell[l]{$\mu$: 0.58 \\ $\sigma^2$: 0.03} & \makecell[l]{$\mu$: 0.60 \\ $\sigma^2$: 0.03} & \makecell[l]{$\mu$: 0.58 \\ $\sigma^2$: 0.04} & \makecell[l]{$\mu$: 0.58 \\ $\sigma^2$: 0.03} \\
GCS & \makecell[l]{$\mu$: 7.29 \\ $\sigma^2$: 12.42} & \makecell[l]{$\mu$: 8.87 \\ $\sigma^2$: 21.54} & \makecell[l]{$\mu$: 7.20 \\ $\sigma^2$: 14.55} & \makecell[l]{$\mu$: 7.60 \\ $\sigma^2$: 13.23} & \makecell[l]{$\mu$: 7.88 \\ $\sigma^2$: 16.47} \\
Urine & \makecell[l]{$\mu$: 89 \\ $\sigma^2$: 9741} & \makecell[l]{$\mu$: 103 \\ $\sigma^2$: 14923} & \makecell[l]{$\mu$: 94 \\ $\sigma^2$: 17197} & \makecell[l]{$\mu$: 71 \\ $\sigma^2$: 8947} & \makecell[l]{$\mu$: 88 \\ $\sigma^2$: 12805} \\
pH & \makecell[l]{$\mu$: 7.37 \\ $\sigma^2$: 0.01} & \makecell[l]{$\mu$: 7.39 \\ $\sigma^2$: 0.00} & \makecell[l]{$\mu$: 7.36 \\ $\sigma^2$: 0.01} & \makecell[l]{$\mu$: 7.34 \\ $\sigma^2$: 0.01} & \makecell[l]{$\mu$: 7.36 \\ $\sigma^2$: 0.01} \\
$PaCO_2$ & \makecell[l]{$\mu$: 34.77 \\ $\sigma^2$: 54.11} & \makecell[l]{$\mu$: 38.53 \\ $\sigma^2$: 29.47} & \makecell[l]{$\mu$: 39.08 \\ $\sigma^2$: 47.43} & \makecell[l]{$\mu$: 36.61 \\ $\sigma^2$: 84.38} & \makecell[l]{$\mu$: 37.62 \\ $\sigma^2$: 55.24} \\
$PaO_2$ & \makecell[l]{$\mu$: 128 \\ $\sigma^2$: 3375} & \makecell[l]{$\mu$: 143 \\ $\sigma^2$: 6336} & \makecell[l]{$\mu$: 129 \\ $\sigma^2$: 3437} & \makecell[l]{$\mu$: 118 \\ $\sigma^2$: 3109} & \makecell[l]{$\mu$: 130 \\ $\sigma^2$: 4465} \\
\bottomrule
\addlinespace
\addlinespace
\end{tabular}
\end{table}

\section{Convergence Diagnostics}
\label{app:D}
\subsection{Convergence Diagnostics on Simulated Data}
\begin{figure}[H]
    \centering
    \includegraphics[scale = 0.235]{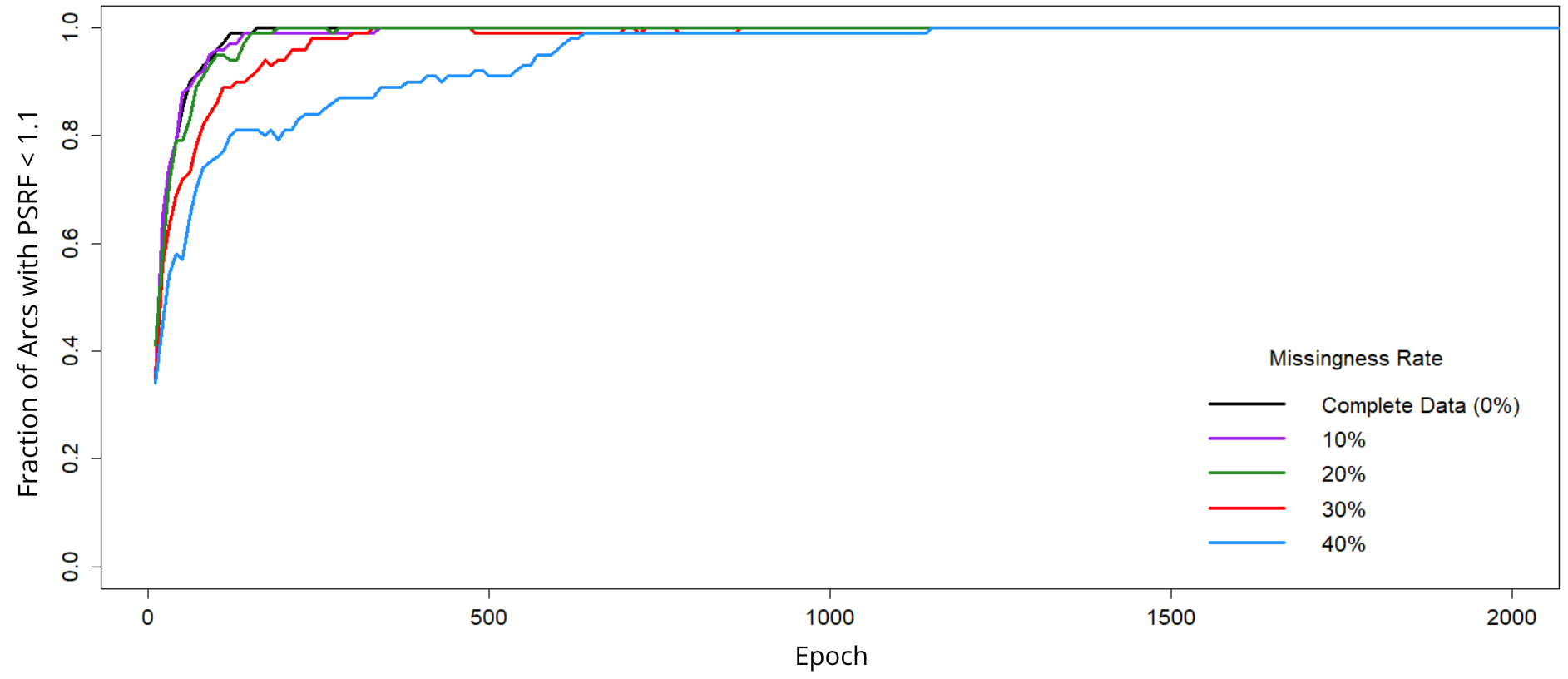}
    \caption{Convergence Diagnostic for Network reconstruction averaged over 5 simulations for simulated datasets with different missingness rates.}
    \label{fig:networkconvergence}
\end{figure}

\begin{figure}[H]
    \centering
    \includegraphics[scale = 0.24]{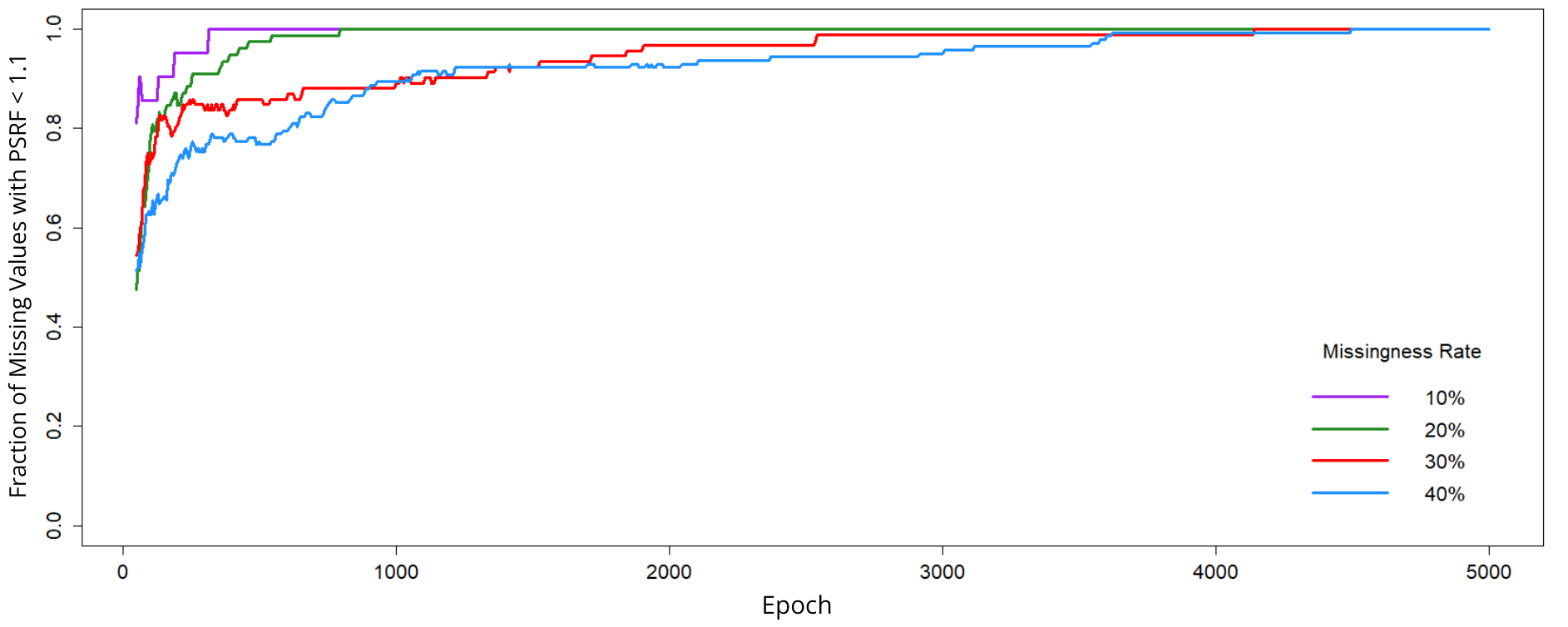}
    \caption{Convergence Diagnostic for Missing Value imputation averaged over 5 simulations for simulated datasets with different missingness rates.}
    \label{fig:missingconvergence}
\end{figure}
\subsection{Convergence Diagnostics on Intensive Care Data}
\begin{figure}[H]
    \centering
    \includegraphics[scale = 0.24]{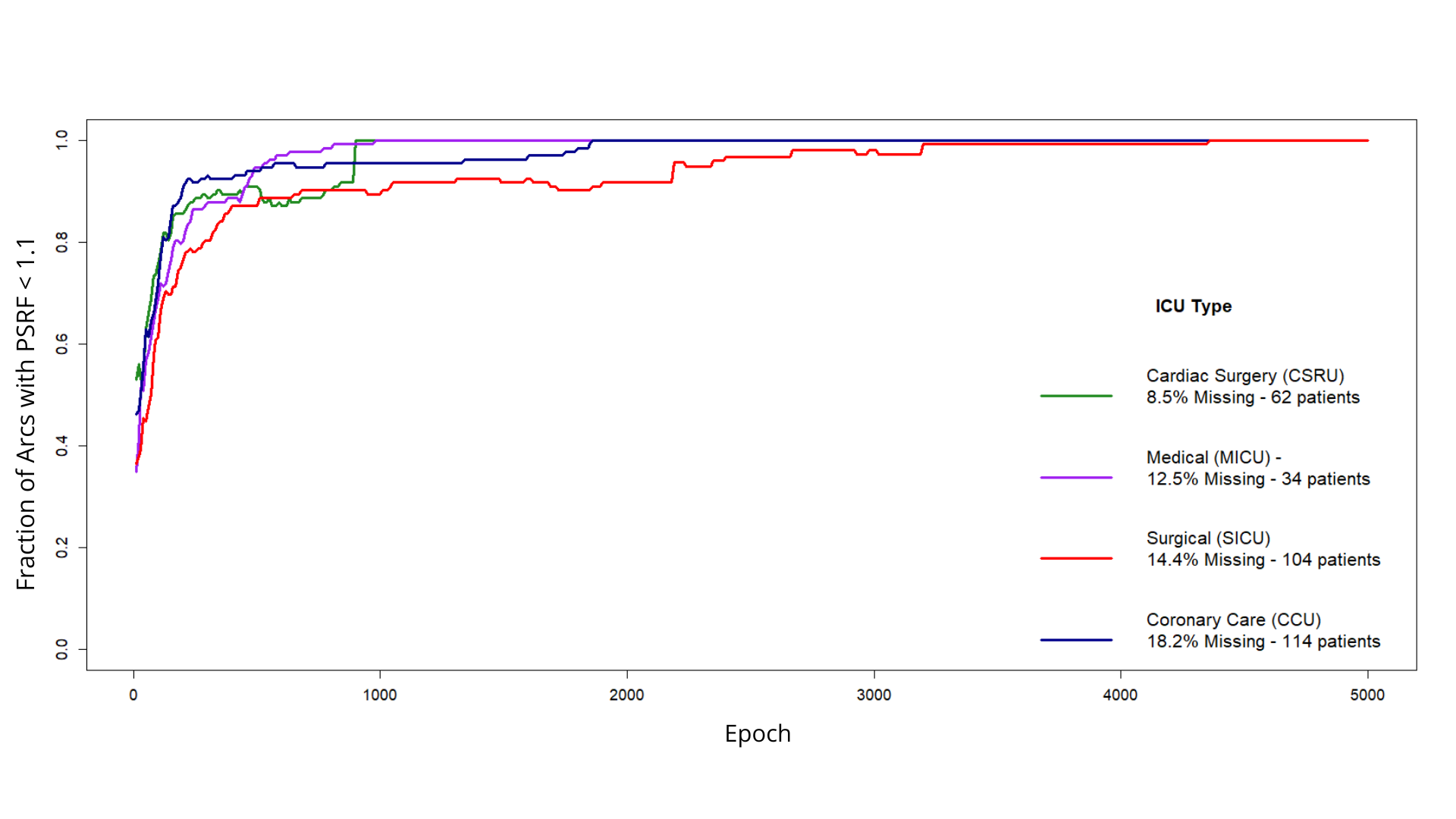}
    \caption{Convergence Diagnostic for Network reconstruction of PhysioNet data segmented per ICU group averaged over 5 simulations.}
    \label{fig:networkconvergence2}
\end{figure}

\begin{figure}[H]
    \centering
    \includegraphics[scale = 0.24]{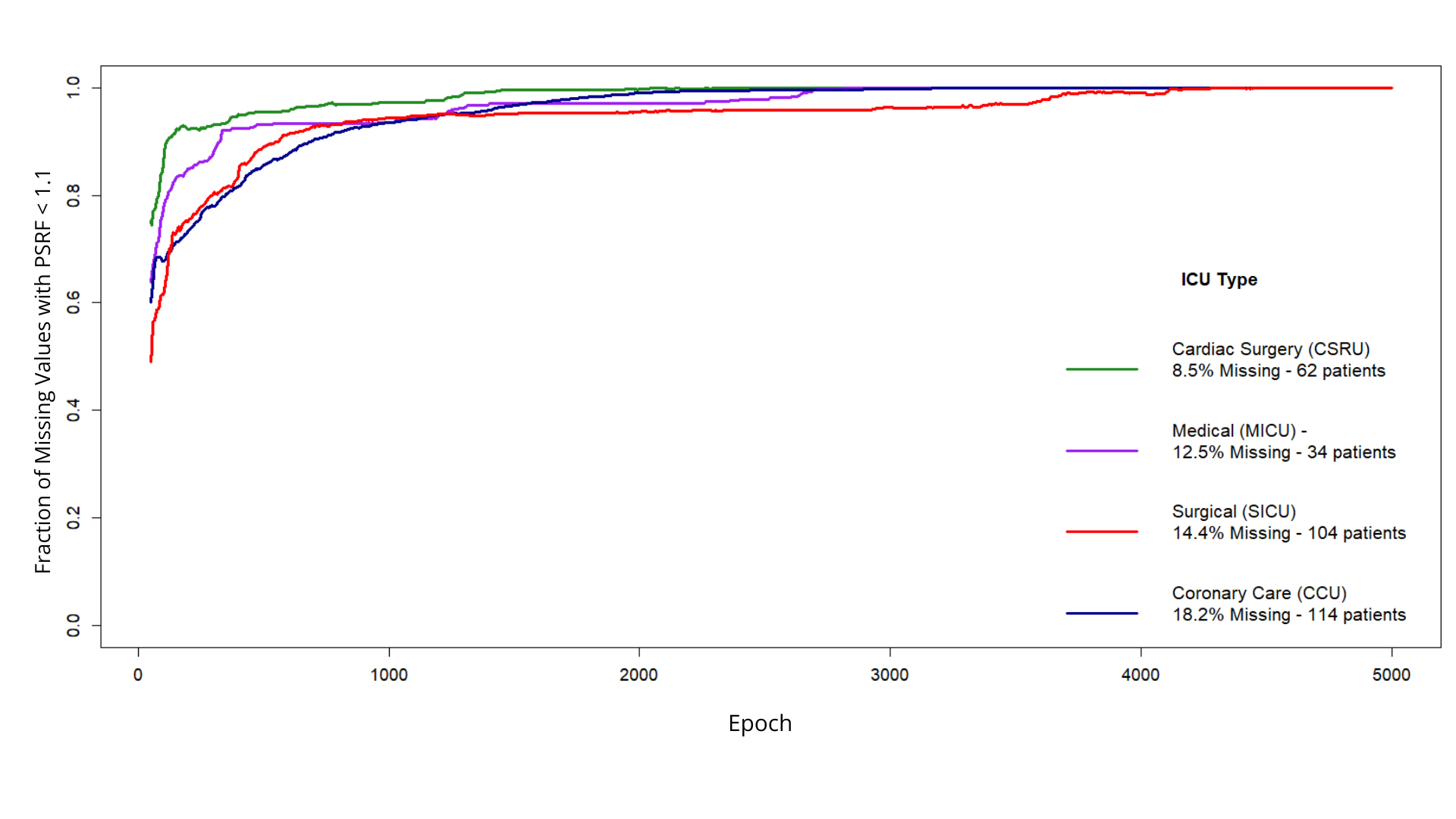}
    \caption{Convergence Diagnostic for Missing Value imputation of PhysioNet data segmented per ICU group averaged over 5 simulations.}
    \label{fig:missingconvergence2}
\end{figure}

\section{Reconstructed Networks}
\label{app:E}
\begin{figure}[H]
    \centering
    \includegraphics[scale = 0.17]{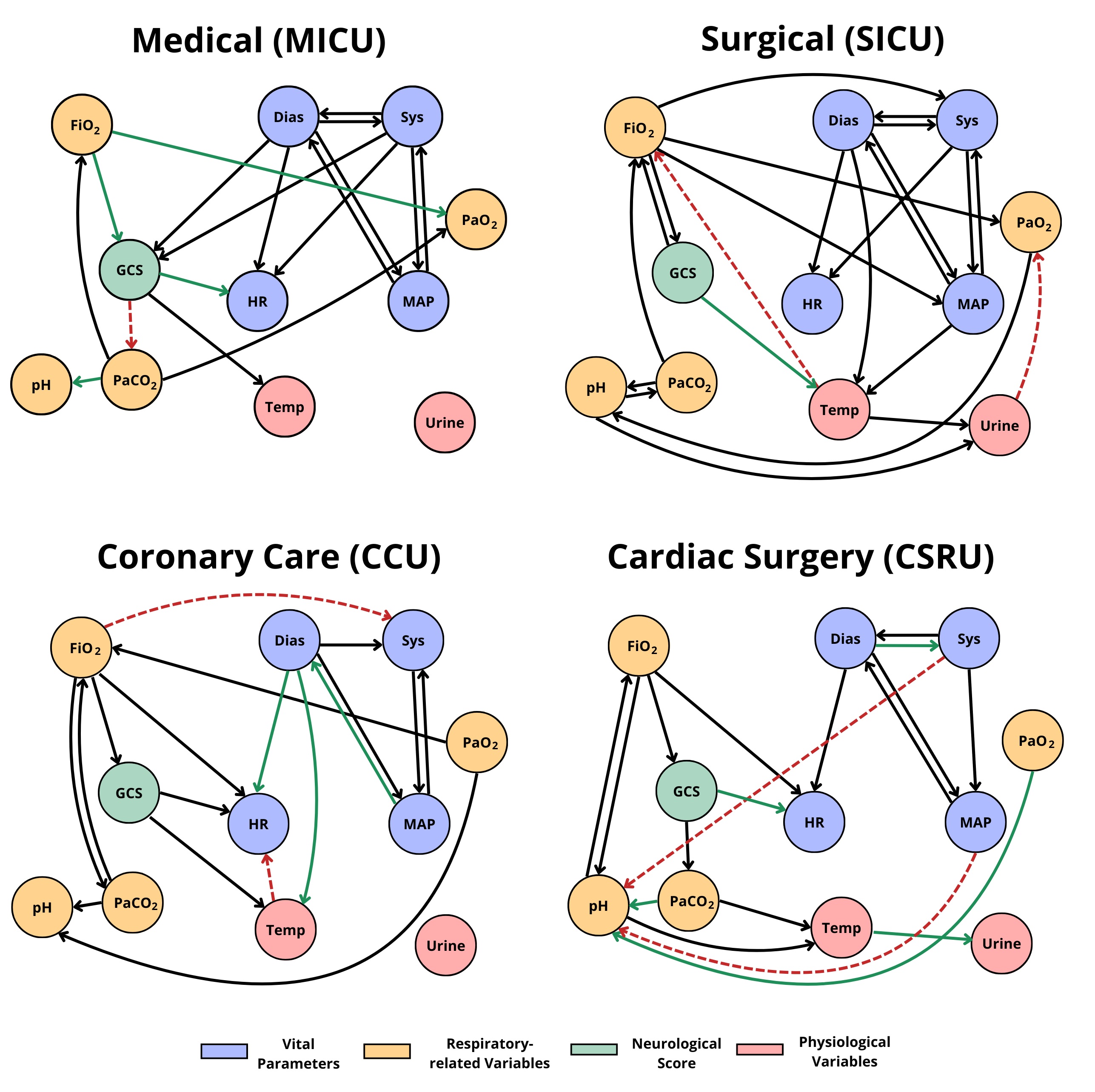}
    \caption{Reconstructed DBNs for each ICU type (MICU, SICU, CCU, CSRU), averaged over five independent simulations for two different data standardization techniques. A threshold of 0.8 is applied to generate the network structures. Arcs are meant to represent temporal relationship with a single temporal lag, namely between nodes at time $t-1$ and nodes at time $t$. Green arcs are present just in the context of global standardization, red dotted arcs are present in local standardization per ICU group only and black arcs are common to the two settings.}
    \label{fig:reconstructednetworks}
\end{figure}

\end{document}